\documentclass[conference]{ieeeconf}
\IEEEoverridecommandlockouts
\usepackage{cite}
\usepackage{amsmath,amssymb,amsfonts, array, arydshln}
\usepackage{algorithmic}
\let\labelindent\relax
\usepackage{enumitem}
\usepackage[hidelinks]{hyperref}
\usepackage{graphicx}
\usepackage{textcomp}
\usepackage{array}

\usepackage{hyperref}
\hypersetup{
    colorlinks=true,
    linkcolor=blue,
    filecolor=magenta,      
    urlcolor=cyan,
    pdftitle={cramp},
    pdfpagemode=FullScreen,
    }

\graphicspath{{./figures/}}
\usepackage{xcolor}
\def\BibTeX{{\rm B\kern-.05em{\sc i\kern-.025em b}\kern-.08em
    T\kern-.1667em\lower.7ex\hbox{E}\kern-.125emX}}
\begin{document}

\title{\textbf{HEROES}: Unreal Engine-based Human and Emergency Robot Operation Education System}

\author{Anav Chaudhary, Kshitij Tiwari and Aniket Bera% 
\\\textit{Department of Computer Science, Purdue University, USA}
\\\textbf{Code}: \href{https://github.com/Anav-117/HEROES}{https://github.com/Anav-117/HEROES}
\thanks{The authors are with the Department of Computer Science, Purdue University, USA,
        {\tt\small \{chaudh75,tiwarik,aniketbera\}@purdue.edu}}
}

\maketitle
\vspace{-10pt}

%% Main paper sections
\begin{abstract}
Training and preparing first responders and humanitarian robots for Mass Casualty Incidents (MCIs) often poses a challenge owing to the lack of realistic and easily accessible test facilities. While such facilities can offer realistic scenarios post an MCI that can serve training and educational purposes for first responders and humanitarian robots, they are often hard to access owing to logistical constraints. To overcome this challenge, we present \textbf{HEROES}- a versatile Unreal Engine-based simulator for designing novel training simulations for humans and emergency robots for such urban search and rescue operations. The proposed HEROES simulator is capable of generating synthetic datasets for machine learning pipelines that are used for training robot navigation. This work addresses the necessity for a comprehensive training platform in the robotics community, ensuring pragmatic and efficient preparation for real-world emergency scenarios. The strengths of our simulator lie in its adaptability, scalability, and ability to facilitate collaboration between robot developers and first responders, fostering synergy in developing effective strategies for search and rescue operations in MCIs. We conducted a preliminary user study with an average score of 8.1 out of 10 supporting the ability of HEROES to generate sufficiently varied environments and a score of 7.8 out of 10 affirming the usefulness of the simulation environment. HEROES has been integrated with ROS and has been used to train an RL model for a real robot as a proof of concept.
\end{abstract}
\section{Introduction}
Mass Casualty Incidents (MCIs) are scenarios that can benefit heavily from the intricate integration of mobile robots in traditional human workflows involving first responders given the scale of such scenarios \cite{Habib2010Robot, Tadokoro2013Disaster}. The nature of such incidents causes emergency service resources- human or equipment- to be overwhelmed by the sheer difference in numbers. These scenarios are often not limited to a difference in victims and service providers, but may also comprise situations where a single interaction requires dedicated human involvement, often derived from a pool of limited human resources. 
\begin{figure}[!htbp]
    \centering
    \includegraphics[scale=0.22]{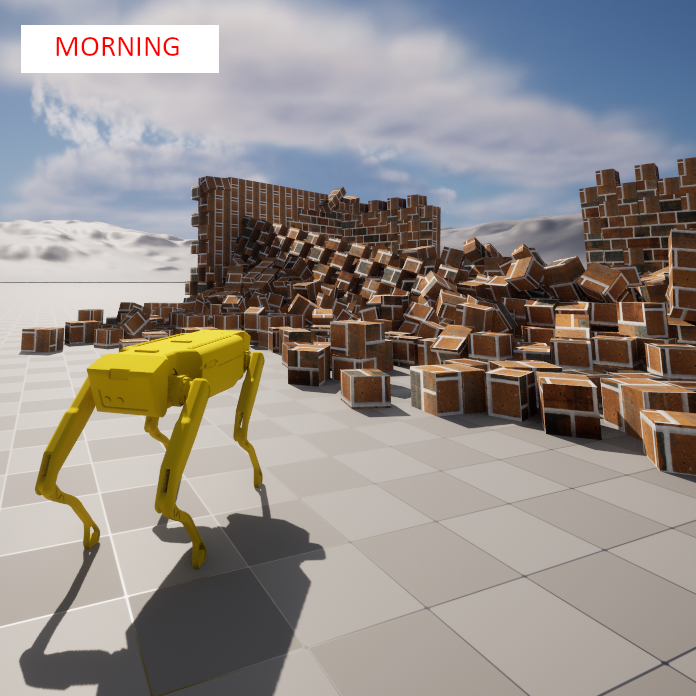}
    \includegraphics[scale=0.22]{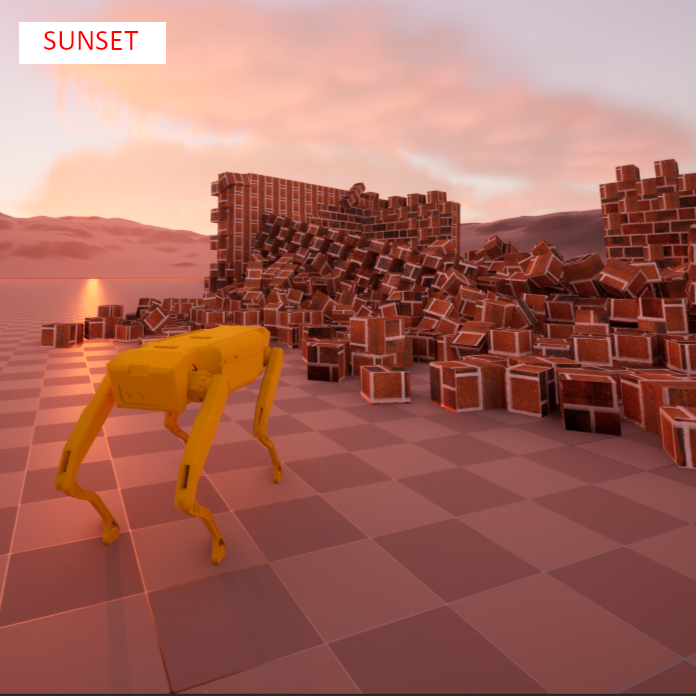} \\
    \vspace{3pt}
    \includegraphics[scale=0.105]{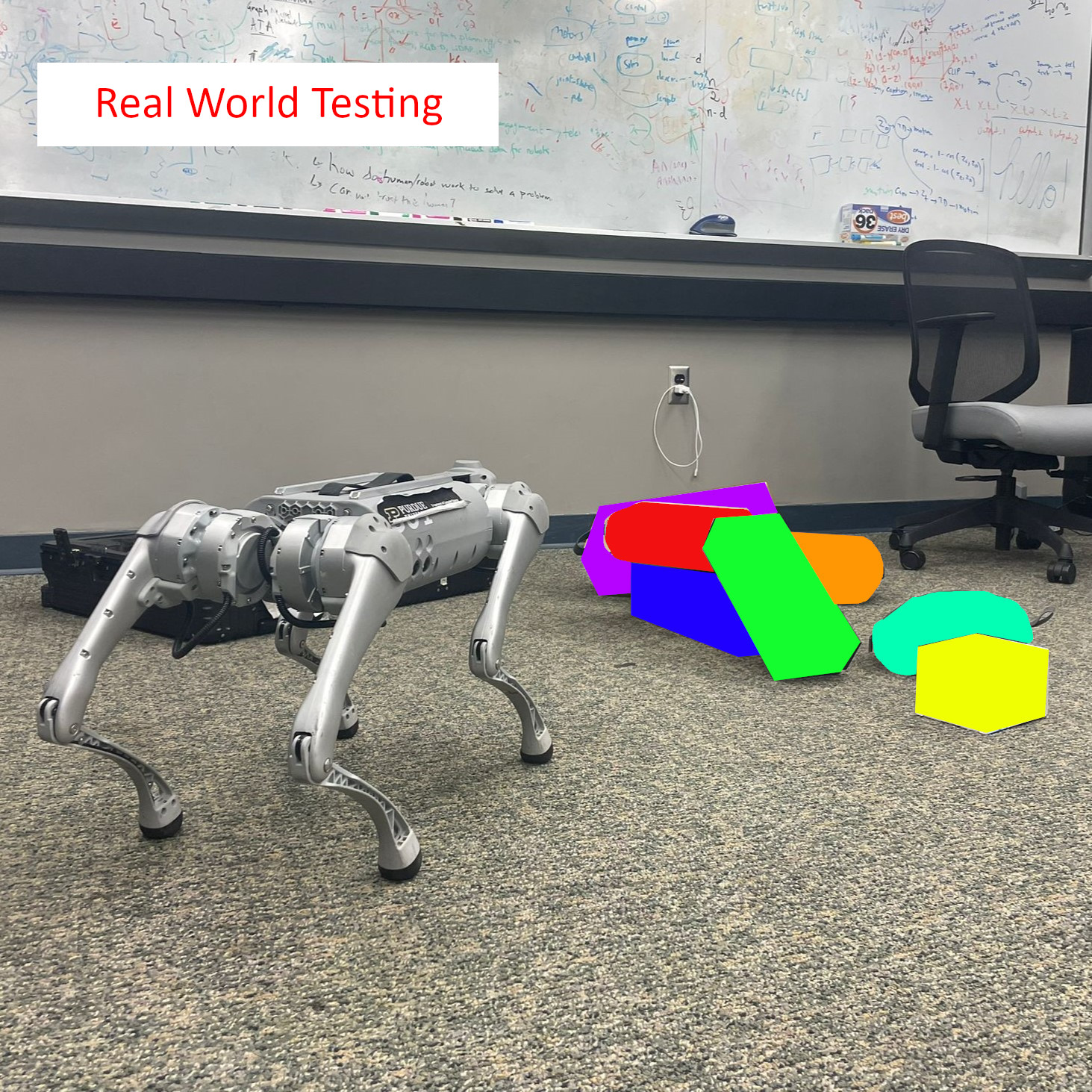}
    \includegraphics[scale=0.22]{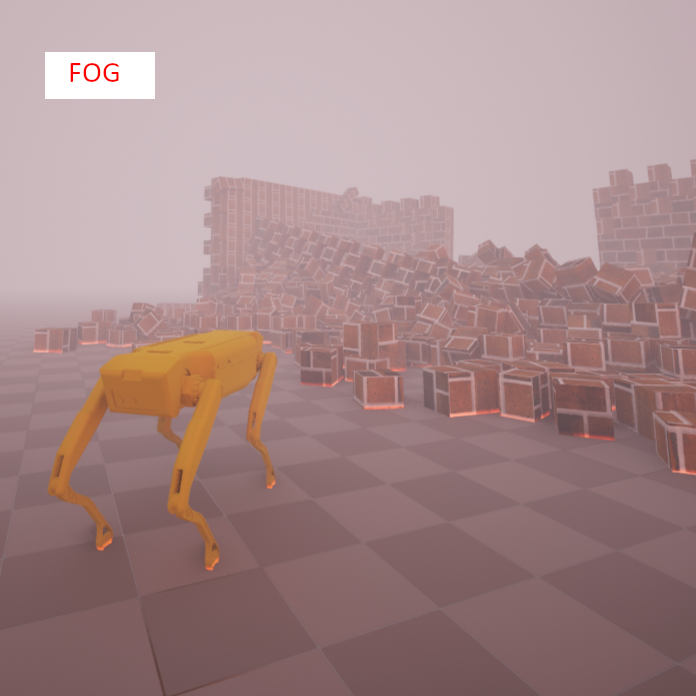} \\
    \vspace{3pt}
    \includegraphics[scale=0.295]{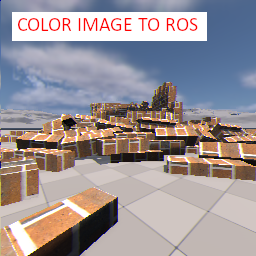}
    \includegraphics[scale=0.295]{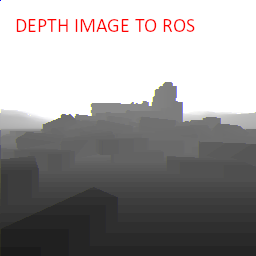}
    \vspace{3pt}
    \includegraphics[scale=0.295]{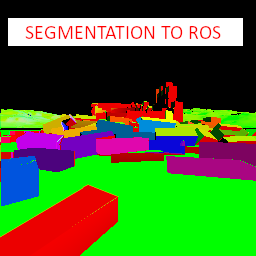} \\
    \caption{\textit{Urban post-MCI destruction environments are not represented during robot navigation training and yet offer unique challenges. The figure showcases the highly irregular environment created by our proposed simulator \textbf{HEROES} with a wide variety of lighting scenarios due to structural collapse post MCIs. In addition, the figure showcases the type of perceptual feedback that can be extracted, ordered from left to right, such as Color Image, Depth, and Semantic Segmentation. Finally, the figure also showcases how the simulator is being used to train robots to tackle similar environments in the real world.}}
    \label{fig:motivation}
\end{figure}
One of the prime examples of this is the concept of Urban Search and Rescue. This activity requires the deployment of a large number of resources to provide emergency services to a possibly smaller group. This creates obstacles in the proper delivery of such services. Furthermore, as the component of the Search yields results, resources must be diverted to the Rescue and treatment of individuals, i.e., primary triage, which diverts resources away from searching for more victims.
Current applications of robot integration in MCI response workflows use a wide variety of robots, such as underwater robots for water disasters, small-sized UGVs (unmanned ground vehicles) deployed for victim search during earthquakes, and UAVs (unmanned aerial vehicles) for monitoring volatile situations such as volcanic eruptions \cite{Tadokoro2013Disaster}. However, such applications often utilize robots trained for navigation on naturally formed terrain \cite{Lee202Learning, Miki2022Learning}, or simulations of such terrain \cite{Bertolino2019Advanced, Tanzi20223D}. MCIs often present a unique set of obstacles in the form of artificial terrain, such as rubble piles and partially demolished buildings, as shown in Fig.~\ref{fig:motivation}. The use of such environments for the purpose of training is done through real-world simulations in facilities such as the Disaster City facility maintained by Texas A\&M Engineering Extension Service (TEEX) \cite{TEEX}.
However, such facilities are expensive to maintain and operate. Furthermore, the number of such facilities is limited, and the simulations offered are also limited in various situations and layouts. Therefore, the current status quo can benefit from a method of computer-aided simulation of such incidents, which can offer large variability, and be both cost-effective and scalable. 
Robot navigation is increasingly making use of machine learning and deep learning pipelines, which require large and diverse datasets to achieve results that translate well to real-world situations. While general computer simulation has been used to generate such datasets in sim-to-real approaches \cite{Zhao2020Sim} and especially for perceptual agents \cite{Mania2019Framework}, MCIs present a new form of environment not yet explored in past works. Rapid and varied generation of such environments can provide additional data for existing machine learning and deep learning pipelines that will allow robots to better adapt to such environments.

We present, \textit{HEROES - an Unreal Engine-based human and emergency robot operation education system}. HEROES is a simulation framework useful for training mobile robots capable of operating in urban MCIs and performing automated search operations through unknown and dynamic environments, as well as training data-intensive machine learning models. To this end, we present the following novel contributions:
\vspace{-2pt}
\begin{itemize}
    \item Traditional methods of navigating challenging terrain used by robots are trained on natural terrain or common urban terrain \cite{Lee202Learning, Miki2022Learning}, which is not an accurate analog for urban terrain affected by MCIs. Such terrain is rife with hazards and unforeseen obstacles and routes due to the presence of \textbf{physical impediments such as rubble}. We present a simulation of such terrain to be used to train robots to navigate such an environment effectively. 
    \item The high variability presented through the simulation provides \textbf{a large number of unique simulation environments} that can be tailored by users to represent different forms of MCIs.
    \item By utilizing a \textbf{physical simulation of urban destruction}, we present a way to generate variations from the same starting environment, thus providing an environment where robots can interact with dynamic objects (rubble pieces that may move during robot interaction).
    \item We also showcase using the generated models from HEROES to train a robot to navigate through complex terrains using Reinforcement Learning.
\end{itemize}
\section{Related Works}
This section discusses different simulation frameworks used in training robotics pipelines and in other interfaces, particularly in MCIs.

\subsection{Training professionals for MCIs through computer-aided simulations}
MCIs, require complex coordination and efficient action by several different entities, such as first responders, medical professionals, and technical specialists. Furthermore, the random and chaotic nature of such incidents often makes it difficult to prepare for every possibility, with service providers facing unexampled scenarios regularly. Current methods for training often lack the realism or variability to model MCIs effectively. Computer-aided simulations, through the use of virtual environments provide an incredible substitute for such training \cite{Pucher2014Virtual, Hsu2013State}. Virtual reality to simulations of MCI environments have been found to be an effective method of training medical professionals and first responders \cite{Farra2012Integrative}. Multiple instances of MCI simulation for training have proven fruitful both in cases of initial training \cite{Cohen2013Emergency, Cohen2012Major, Heinrichs2010Training}, and in the assessment of preparedness \cite{Cohen2013Tactical}.
These works have highlighted the efficacy of computer simulations in training for MCIs and thus showcase that virtual simulations of such events can represent actual events accurately. 

\subsection{Using computer simulations to train robot navigation}
Simulated training of robots has now emerged as a pragmatic and structured method of robot training. Sim-to-real transfer can be used effectively to train different kinds of robotic systems in various environments, with promising results. Computer simulation environments are proving to increase the effectiveness of training by providing a large amount of data points (potentially infinite), and also delivering a safe environment for training \cite{Zhao2020Sim}. Randomization techniques are employed widely to generate large amounts of unique and varied environments in an attempt to produce better results in unforeseen scenarios \cite{Tobin2017Domain, Peng2018Sim}. Xie et al. explore the use of dynamics randomization in robot training and conclude that without the use of on-robot adaptation, dynamics randomization may not always be sufficient for the transfer of learned behavior to the real world \cite{Xie2021Dynamics}. This establishes the need for learning adaptive behavior over unstable terrain where sudden forces may be encountered, such as unstable rubble piles. In other works, such as Gangapurwala et al. \cite{Gangapurwala2022RLOC}, the use of reinforcement learning to achieve dynamic locomotion over uneven terrain provides good results over natural terrain but lacks the unnatural terrain deformities often present post-MCIs. Training of perceptual agents is increasingly being conducted through deep learning methods wherein researchers are attempting to improve robot navigation through domain adaptation for visual control \cite{Zhang2019VR}, learning semantic details about the environment \cite{Sadeghi2019Divis}, or by using real-world executions to improve simulation parameters \cite{Chebotar2019Closing}, where each approach benefits from synthetic datasets. However, these approaches tend to focus on common real-world environments and use datasets that may not translate to a post-MCI environment. 
These works highlight the effectiveness of synthetic datasets in the training of robot navigation but also expose the gap between currently available datasets, which provide accurate representations of normal real-world scenarios but are unable to represent scenarios where such environments may be exaggeratedly deformed.

\subsection{Use of simulation in training robot navigation for MCI environments}
MCIs often present robots with challenging tasks. These include navigation in new and complex environments using semantic cues not present during training \cite{Sadeghi2019Divis}. However, such navigational cues and references are not readily available in MCIs. Instead, the robots may be required to navigate through unconventional routes, where human traversal is not possible \cite{Bertolino2019Advanced}. This presents a form of the \emph{reality gap} problem, however, in this case, the \emph{reality gap} emerges from the difference in the simulated environment and the targeted use environment of the robot. Additionally, robots will often need to traverse challenging and highly uneven terrain. Simulations have been used to train such behavior quite successfully in different kinds of robots, from quadrupedal \cite{Lee202Learning, Miki2022Learning}, and bipedal \cite{Yu2019Sim} to rovers \cite{Bertolino2019Advanced}. However, these simulations again present a similar limitation, as MCI terrain may differ wildly from challenging terrain found in everyday conditions. 
In addition to these, it is imperative to be absolutely aware of the limitations of the robots used and the expectations from the same \cite{Tanzi20223D, Tanzi2020Autonomous, Tiwari2019Unified, tiwari2018estimating}. In disastrous situations, an unexpected failure of the robot can cause a loss of vital onsite time and resources, as well as a general loss of resource availability in the future. Simulation provides a safe environment to explore the limitations and expectations of robotic systems in various MCIs \cite{Tanzi2020Autonomous} and different kinds of robots in similar environments, allowing both the exploration of a single robot's limitations, as well as the comparative analysis of multiple robots.
These publications shed light on the obstacles faced by robots during the navigation of post-MCI environments, thus providing insight into how MCI simulations can be used to address such issues.

These past works show the current research and applications of robot navigation in unstable and uneven terrain and have highlighted the need for vast and varied synthetic data pertaining to highly specific situations that may not be adequately represented by real-world datasets or synthetic datasets modeled after normal real-world scenarios. Furthermore, they highlight how realistic, physics-inspired simulations may accurately provide situations close to MCIs that are difficult to simulate in the real world and how datasets acquired from such simulations may help address additional problems noted in robot deployment during MCIs.
\section{HEROES: Post-MCI Environment Simulator}
In the following sections, we describe the HEROES simulator framework in detail. First, we define and describe the individual destructible unit of the simulator (Section~\ref{subsec:indiv-unit}). Then, we explain the process of setting up the simulation environment (Section~\ref{subsec:sim-env}). Finally, we describe the destruction events and other possible interactions used to create a variety of scenarios in the HEROES simulator (Section~\ref{subsec:destr-events}). 

\subsection{Individual destructible unit} \label{subsec:indiv-unit}
The simulation makes use of the \emph{Chaos Physics} \cite{UnrealEngineChaosDestruction} to fracture meshes and simulate physical interactions of the rubble pieces with the environment. However, at present, the capabilities of the \emph{Chaos Physics} system do not allow the ability to fracture meshes at run-time. This excludes the possibility to use large-scale static meshes that are created during use. Therefore, to overcome this drawback, we define an \emph{Individual destructible unit} to serve as a predefined and pre-fractured mesh, using which users can construct large-scale environments.
\begin{figure}[!htbp]
    \centering
    \includegraphics[scale=0.1]{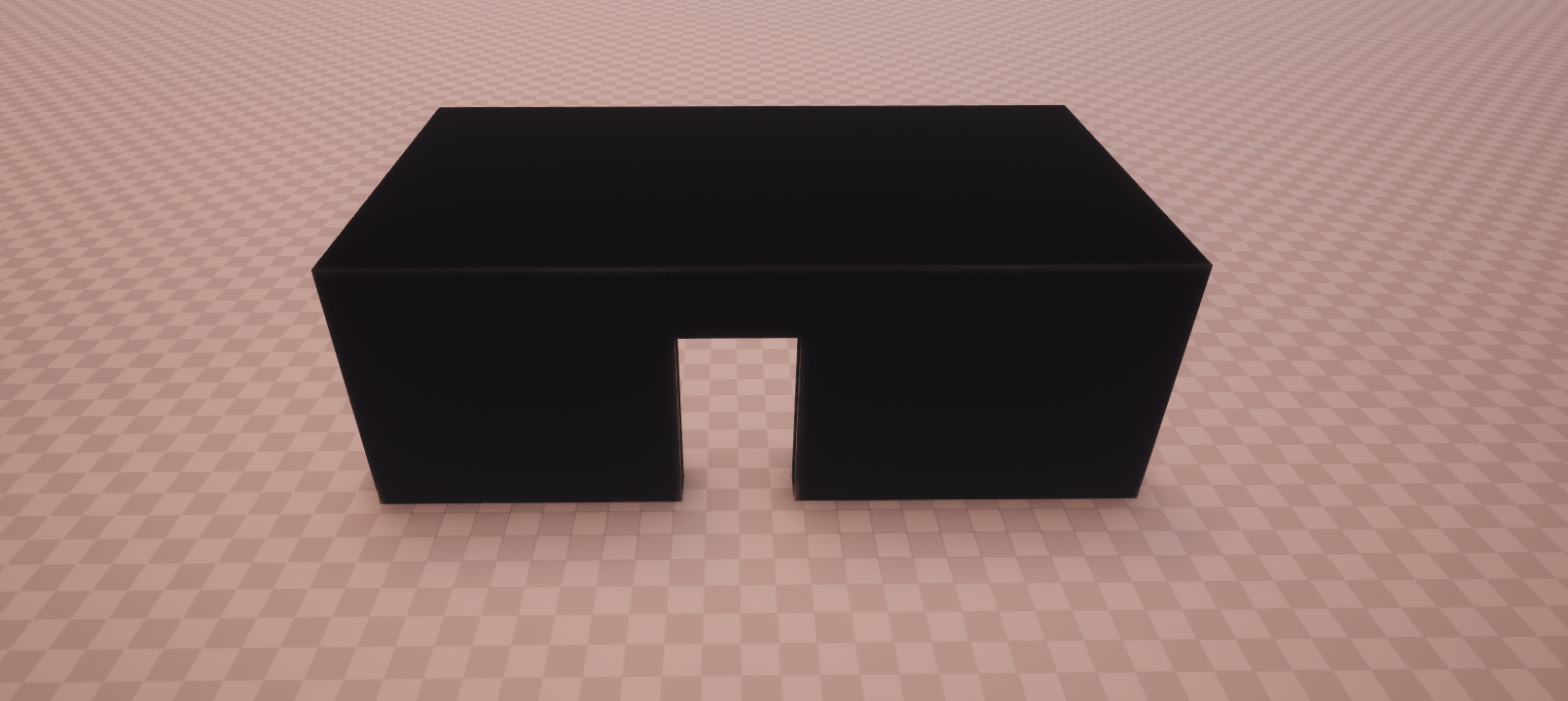}
    \includegraphics[scale=0.2]{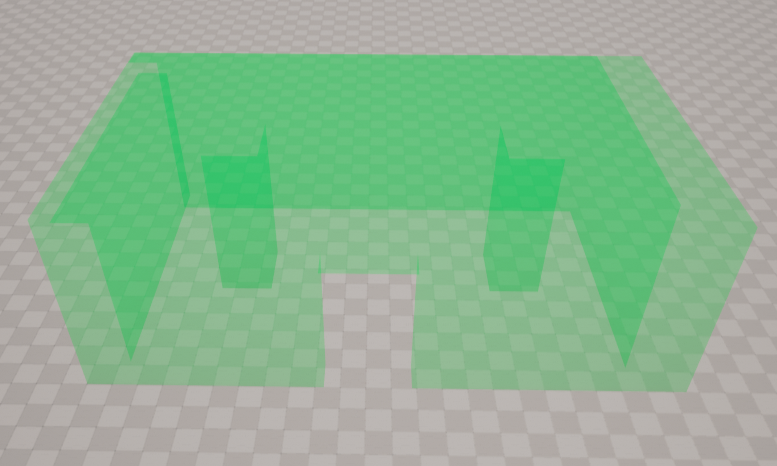} \\
    \vspace{3pt}
    \includegraphics[scale=0.104]{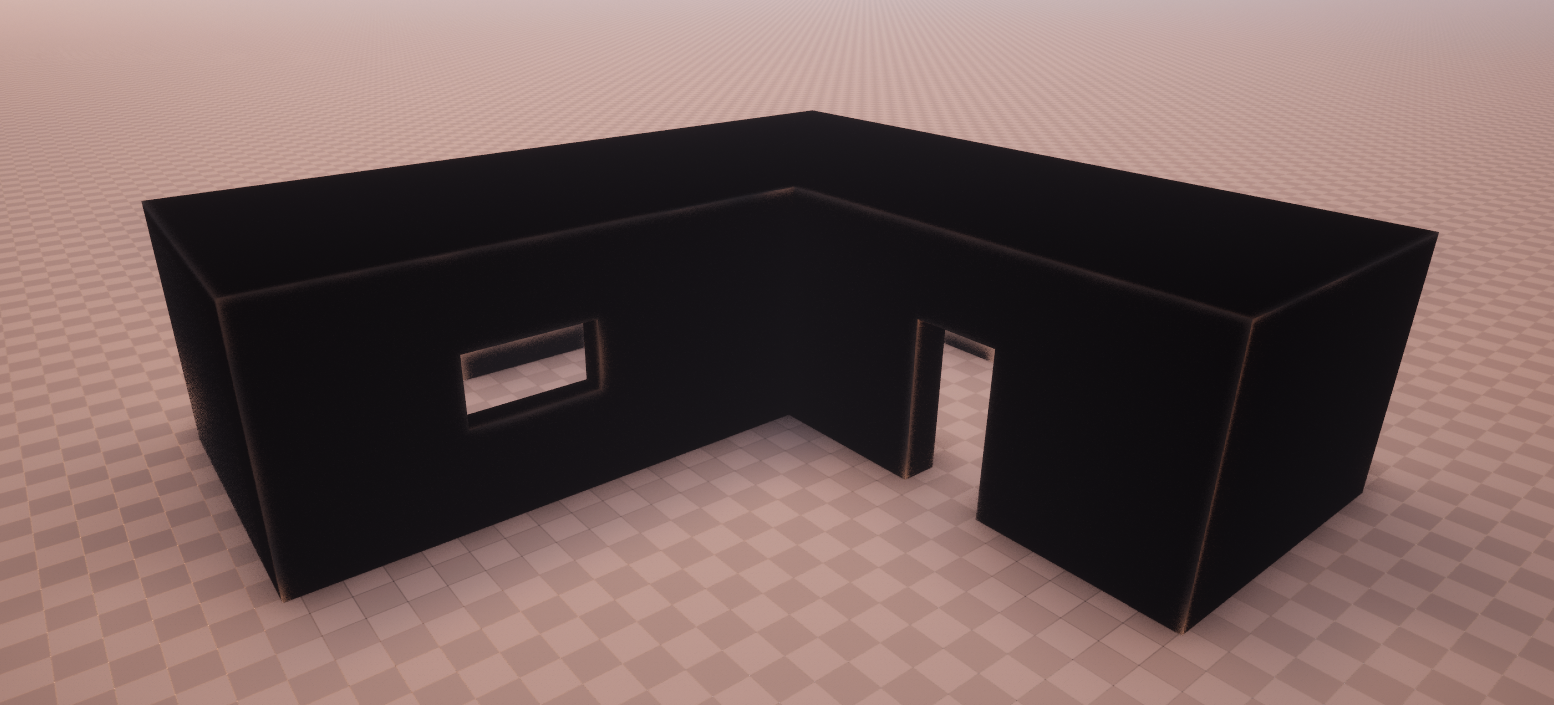}
    \includegraphics[scale=0.2]{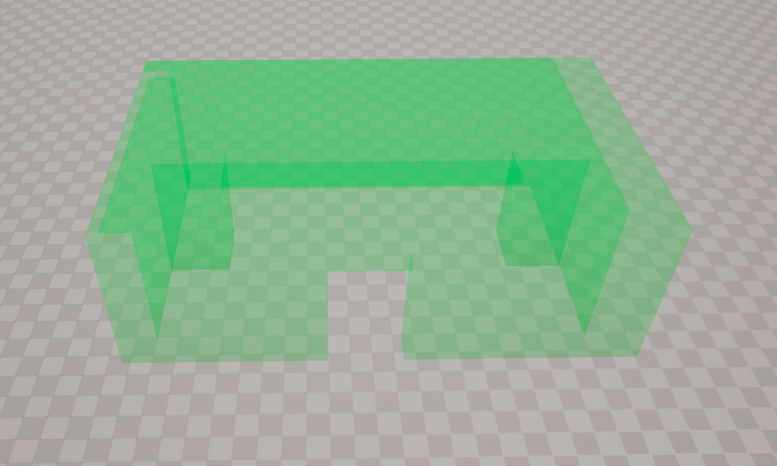}
    \caption{\textit{Multiple different types of Rooms can be used to construct the simulation environment. In a clockwise order, a simple Room with 1 doorway, an L-shaped Room with 1 doorway and 1 window, and translucent views of a 1 doorway room with 2 pillars for support, and a 1 doorway room with a beam and two wall supports are shown.}}
    \label{fig:room-types}
\end{figure}
The aforementioned unit, henceforth named the \emph{Room}, is defined as a single predefined and pre-fractured static mesh, which is subjected to the physics simulation. A large set of Rooms are designed using 3D modeling software, such as Blender, to serve as the building blocks of the simulation environment. Such meshes are fractured using Unreal Engine's in-engine fracture editor before runtime, and every iteration of the simulation will result in the same fracture pattern; however, by simulating physical forces during and after fracturing, we are able to achieve randomness in the spread of broken rubble pieces. Using pre-fractured meshes also allows us to account for different material strengths, such as brick and wood. The fracturing is generally done using Voronoi fracturing, where each individual fractured component is defined as a Voronoi cell. The Equation \(V_i = \{x \in X\;|\;  d(P_i, x) \leq d(P_j, x) \;\forall\;i\neq j\}\) defines the Voronoi region \(V_i\) using control points (or Voronoi sites) denoted by \(P_i\), as the set of all points \(x\) in the 3-dimensional space $\mathbb{R}^3$ that are closer to the Voronoi site \(P_i\) than they are to any other sites. Voronoi sites are chosen randomly during fracturing, and their number can be altered to provide higher granularity when the room is destroyed. The fracturing pattern can also be defined using randomly perturbed planes to specify specific major fracture lines. A combination of using planes for major fracture lines and Voronoi fracturing for smaller chunks of rubble was used in the experiments specified in this paper.
Each Room in the set is designed to mimic different construction methods and structures, as can be seen in Fig.~\ref{fig:room-types}. Furthermore, using a pre-fractured mesh of the room allows us to more closely simulate the common fracture patterns and points found in different structures. During the fracturing process, the meshes of each Room are broken into several individual meshes, which create a single \emph{Geometry Collection}, and each individual mesh is connected to its neighbors. The connections between pieces are subject to a \emph{Damage Threshold} which aims to replicate the strength of different structures. During the physics simulation, a strain is applied to each connection to determine breakage and necessary forces are used to simulate the MCI. 

\subsection{Simulation environment}\label{subsec:sim-env}
The Simulation environment is a collection of Rooms subjected to physics simulation. This collection of Rooms will serve as the obstacle environment to be used during the training of robot navigation. The simulation environment can affect the behavior of the simulation by allowing Rooms to physically interact with one another during destruction.
Users are provided with a variety of different room configurations in order to represent a variety of construction methods, through which users can construct unique and varied simulation environments. Each Room is placed on a grid, allowing easy positioning through discrete placement positions. Furthermore, each Room can be rotated in increments of 90$^{\circ}$. 
The simulation environment can further be customized to utilize a wide variety of building materials such as brick, concrete, and wood, all of which have separate fracture patterns. In addition, users can also specify environmental parameters such as weather effects and time of day to further increase the variety of the generated data. Table~\ref{table:spec} shows all the possible configuration parameters and exportable data available in the simulator. 
Following the construction of the Simulation Environment, the users can introduce different MCI events by placing them around the environment and tailoring the parameters to their needs. 
% \begin{table}[!htbp]
%  % \setlength\extrarowheight{-3pt}
 
%     \begin{tabular}{ | m{9em} | m{20em} | }
%         \hline
%          \textbf{\textit{\large Configurations}} & \textbf{\textit{\large Available Options}} \\
%          \hline
%          & \\
%          \textbf{Environment} & \\
%          Weather Effects & Fog, Rain, Sunshine \\ 
%          Time of Day & Color, Intensity and Shadow Variations \\
%          &\\
%          \hline
%          &\\
%          \textbf{Building} & \\
%          Fracture Patterns & Uniform Voronoi, Brick, Planar \\
%          Material & Brick, Concrete, Wood \\
%          &\\
%          \hline
%          &\\
%          \textbf{Destruction} & \\
%          Types & Universal Strain, Explosion, Strain Buildup \\
%          Configurations & Linear and squared falloff, Radial bounds of effect\\
%          &\\
%          \hline &\\
%          \textbf{Data Export} & \\
%          Types & Color Image, Depth Field, Segmentation \\
%          & \\
%     \hline
%     \end{tabular}
%     \linebreak

% \end{table}
\begin{table}[!htbp]
\begin{tabular}{|l|l|}
\hline
\textit{\textbf{Configurations}}                                                    & \textit{\textbf{Available Options}}                                                                                                       \\ \hline
\begin{tabular}[c]{@{}l@{}}\textbf{Environment}\\ Weather Effects\\ Time of Day\end{tabular} & \begin{tabular}[c]{@{}l@{}}\\Fog, Rain, Sunshine\\ Color, Intensity and Shadow Variations\end{tabular}                                      \\ \hline
\begin{tabular}[c]{@{}l@{}}\textbf{Building}\\ Fracture Patterns\\ Material\end{tabular}     & \begin{tabular}[c]{@{}l@{}}\\Uniform Voronoi, Brick, Planar\\ Brick, Concrete, Wood\end{tabular}                                            \\ \hline
\begin{tabular}[c]{@{}l@{}}\textbf{Destruction}\\ Types\\ Configurations\end{tabular}        & \begin{tabular}[c]{@{}l@{}}\\Universal Strain, Explosion, Strain Buildup\\ Linear and squared falloff, Radial bounds of effect\end{tabular} \\ \hline
\begin{tabular}[c]{@{}l@{}}\textbf{Data Export}\\ Types\end{tabular}                         & \begin{tabular}[c]{@{}l@{}}\\Color Image, Depth Field, Segmentation\end{tabular}                                 \\ \hline
\end{tabular}
    \caption{List of all currently available configurations using which users can control the simulation environment as well as the simulator parameters and export data options.}
    \label{table:spec}
\end{table}
\subsection{Destruction events}\label{subsec:destr-events}
\begin{figure*}[!htp]
    \centering
    \includegraphics[clip=true,trim={2cm 0 2cm 0},height=3cm]{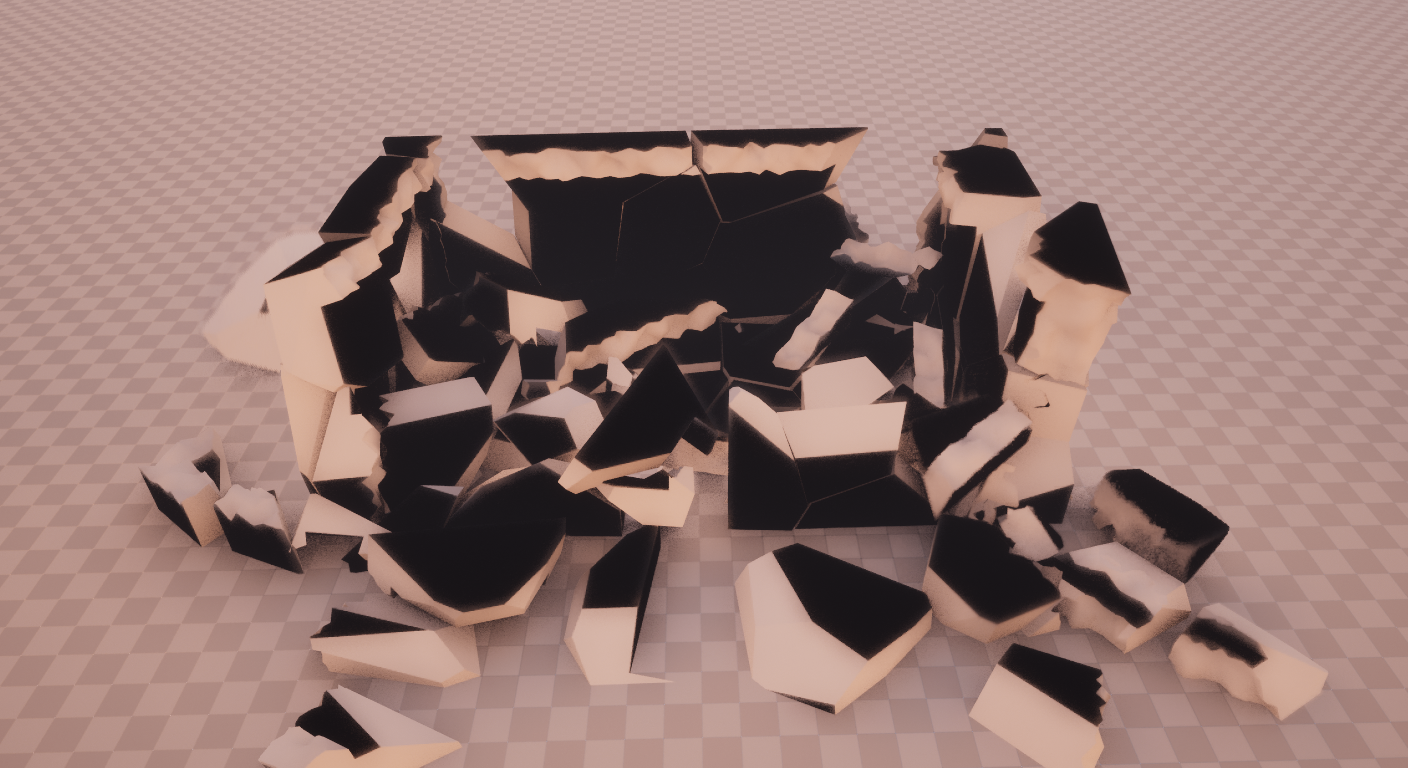}
    \includegraphics[clip=true,trim={2cm 0 2cm 0},height=3cm]{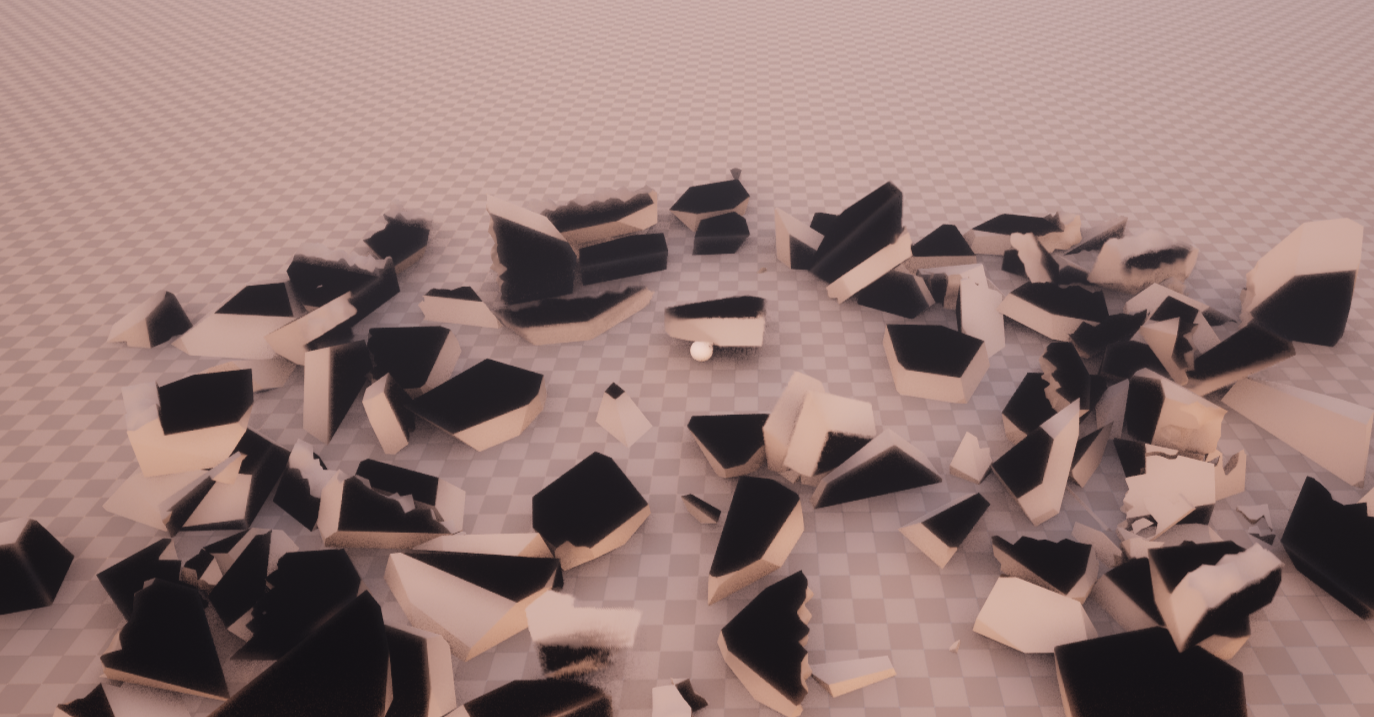}
    \includegraphics[clip=true,trim={2cm 0 3cm 0},height=3cm]{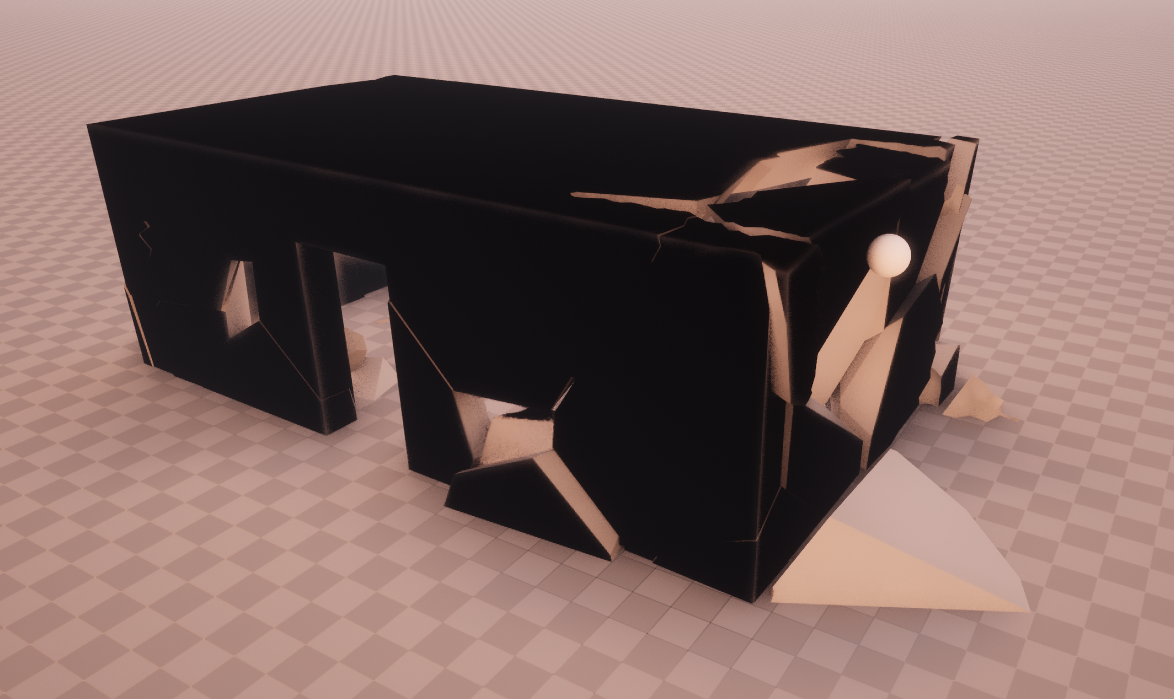}
    \caption{\textit{Different forms of destruction events can be generated via the HEROES simulator. Left: Earthquake-sourced strain, Middle: damage due to explosion, Right: constrained building collapse.}}
    \label{fig:destruction}
\end{figure*}
A Destruction event in the simulation is an event representative of an MCI. These events can be placed in the simulation environment and will be responsible for replicating the MCI in the simulation. Each fractured component of a room is connected to its neighboring fractured components through joints \(x\), with each joint having a strain threshold of \(T_s\). If any joint of a fractured component is subject to a strain higher than \(T_s\), all joints of said fracture component are broken and it acts as an independent physics object. There are 3 Destruction events modeled in the simulation with provisions to add more, as showcased in Fig.~\ref{fig:destruction}. The available events are:
\begin{itemize}[leftmargin=*]
    \item \textbf{Building collapse} triggered through earthquake-sourced strain. This event seeks to replicate the collapse of buildings during an earthquake and the subsequent alteration of the navigation environment. Uniform strain is applied to all pieces of each Room's Geometry Collection throughout the Simulation Environment. The Equation \(S_x = \{M\;|\;x \in X\}\) defines the strain applied on each joint \(S_x\). The strain is calculated uniformly for each joint \(x\) in the set of all joints \(X\) and the magnitude of the strain \(M\) remains constant.
    \item \textbf{Explosive Collapse}. This event replicates the damage caused by an explosion, with large amounts of strain applied to certain areas of the simulation environment. Furthermore, a linear velocity is also provided for all pieces of rubble generated during the application of strain. This helps model the outward dispersion of rubble that can damage the area around the source of the explosion. The equation \(S_x = \{M_s\;|\;x \in R\})\) defines the calculation of the strain applied to a joint \(S_x\) having variable magnitude \(M_s / d^2\) where \(M_s\) is the magnitude at the center of the explosion and \(d\) is the distance between the joint and the center of the explosion, for every joint within a culling region \(R\).
    Following the application of strain, an outward force vector is applied to every component as defined as \(\parallel \Vec{F_x} \parallel = \{M_f\;|\;S_x \geq T_s\; and\; x \in R\}\)  where \(M_f\) is the magnitude of the force applied, and the direction of the force is defined as \(\hat{F_x} = {(\Vec{P_x} - \Vec{C})} \slash {\parallel \Vec{P_x} - \Vec{C} \parallel}\) where \(P_x\) is the 3D position vector of the object whose joint is under consideration and \(C\) is the position vector of the center of the explosion.
    \item \textbf{Constrained building collapse} represents the collapse of certain sections of a building. Strain is applied similarly to earthquake events but is constrained to a smaller region. This event allows only certain areas of the simulation environment to be affected, allowing a simulation of the transition from normal to affected environments. Furthermore, strain is applied incrementally over a duration allowing users to simulate building with different strain thresholds \(T_x\) in the same environment. The equation \(S_x = \{\sum_{t=0}^{t_f} M\;|\;x \in R \}\) defines the calculation of strain over a time \(t_f\).
These Destruction events can further be tailored to suit the user's needs and can be placed around the simulation environment to provide variety without changing the layout of the environment.
\end{itemize}
\subsection{ROS integration}
\begin{figure}[!htbp]
    \centering
    \includegraphics[scale=0.3]{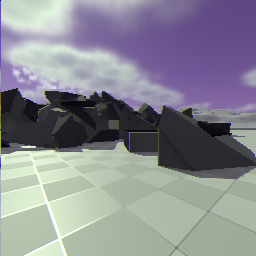}
    \includegraphics[scale=0.3]{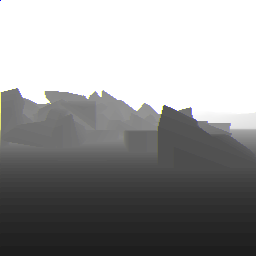}
    \includegraphics[scale=0.3]{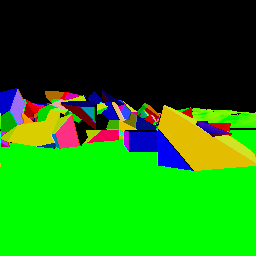} \\
    \vspace{2.5pt}
    \includegraphics[scale=0.3]{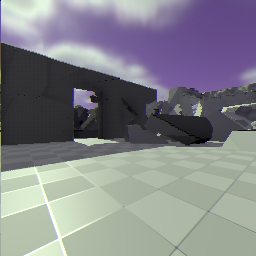}
    \includegraphics[scale=0.3]{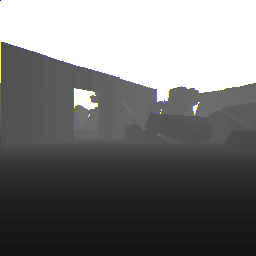}
    \includegraphics[scale=0.3]{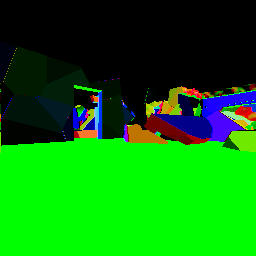} \\
    \vspace{2.5pt}
    \includegraphics[scale=0.3]{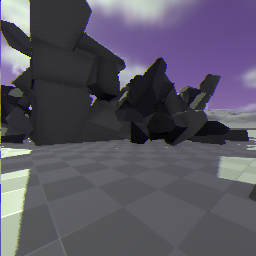}
    \includegraphics[scale=0.3]{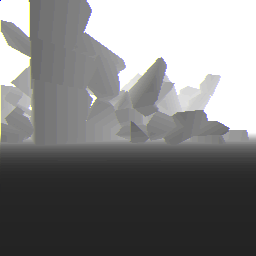}
    \includegraphics[scale=0.3]{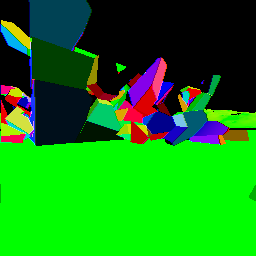}
    \caption{\textit{The ROS Integration allows us to transfer data between the simulator and ROS, which is one of the most popular software development platforms for robotics. The images above show the different forms of data that can be exported from the simulator as images of a quadrupedal robot's point of view. They are arranged as follows; Left: Color Image, Middle: Depth Image, Right: Semantic Segmentation.}}
    \label{fig:ros-integration}
\end{figure}
To make the simulation compatible with current state-of-the-art robot simulation platforms such as ROS, we have added the provision of controlling the robots in the simulation using ROS through a ROSBridge. Using ROS message-passing to control components inside the simulation, defined inside Unreal Engine as a set of physical meshes connected by joints, the robot can be controlled inside the simulation. This will help us translate robot behavior to the physical simulation, allowing them to interact with the physics objects (rubble) in the simulator. Such a scenario can help us address the issue of robot navigation over dynamic terrain (shifting rubble piles as an example) that is not easily replicable using normal real-world environments. 
Users can also export data from the simulator to ROS, including but not limited to images, as seen in Fig.~\ref{fig:ros-integration} from the robot's point of view. This two-way integration of data defines a complete integration of the robot in the simulator with real-world transferability. Additional sensors can be simulated to increase the possible interactions a robot may have with the environment. 
Apart from integration with ROS, the simulator can also be deployed using mixed-reality devices to help train first responders in MCI situations.
% \subsection{Reinforcement Learning}
% Using the ROS integration, we can implement reinforced learning algorithms on robots to help them learn navigation methods in a post-MCI environment. Here, the environment can help provide feedback to the robot in the form of physical collisions with the rubble piles. The robot, required to navigate through the post-MCI landscape can then use this feedback to learn navigation methods over such a terrain in real life. As shown in Fig.~\ref{fig:sim2real}, the robot can now be equipped to effectively handle the terrain of a post-MCI environment.

% The reinforcement learning algorithm is implemented over an environment observed using an observation model, which observes the generated environment from the simulator, using which the algorithm will determine the next state and return a reward \(r_t+1\). This can be modeled as a Markov Decision Process as outlined by Miki et. al. \cite{Miki2022Learning}. Furthermore, following Miki et. al., we can model the policy \(\pi^*\) to maximize the expected reward over the generated terrain as outlined in Eq. \eqref{eq:reward}.
% \begin{equation}\label{eq:reward}
%     \pi^* = \underset{a}{argmax}(E[\sum_{t=0}^{\infty}\gamma^tr_t])
% \end{equation}
\section{Experiments}
To evaluate HEROES's usefulness, we set up a few tasks for beta testers to perform, which will be used to evaluate its ease of use and adaptability. The simulator was tested on a system operating with an Intel i5-10300H CPU and an NVIDIA GeForce GTX1650 Ti GPU.

\subsection{Experiment tasks}
The simulator was tested for usability and effectiveness of integration into existing workflows for robot navigation training. To test the same, a set of beta testers was asked to operate the simulation and perform the following tasks:
\begin{enumerate}[label=(\textbf{Task \arabic*}),align=left,itemsep=3pt,leftmargin=*]
    \item \textit{The reviewers were provided with layout details of an environment and were then asked to recreate the environment as closely as possible.}
    \item \textit{The reviewers were tasked with creating an environment of their choosing.}
\end{enumerate}
The above-mentioned tasks were set to evaluate the ease of reproducibility of an environment, the ease of creating dense, detailed environments, and the ease of creating unique environments from a user's imagination, respectively. The users were provided with only a minimal set of instructions on the controls of the simulator.

The overall ability of the simulator was then evaluated through a questionnaire filled out by all participants. The results of the said questionnaire are highlighted in Table~\ref{table:task-eval} and expounded in the next section. In addition, users also filled out a supplementary questionnaire targeted at quantifying general usability and user feedback. This supplementary questionnaire is outlined in Table~\ref{table:usability-eval}. All answers were collected using a 10-point linear scale.

\subsection{Usability survey}
The simulator's usability was tested using the previously mentioned experiments by a group of 10 beta testers, and the results were collated using questionnaires to judge the effectiveness of the simulator in two ways. The first questionnaire targeted the user's ability to complete the tasks put forward. The results for this are presented in Table~\ref{table:task-eval}. The second questionnaire aimed to understand the general usability of the simulator and the effectiveness that such a tool can have on existing robot training workflows, the results of which are presented in Table~\ref{table:usability-eval}.
\begin{table}[!htbp]
\begin{tabular}{|l|l|}
\hline
\textit{\textbf{Tasks}}                                              & \textit{\textbf{\begin{tabular}[c]{@{}l@{}}Average \\ Score\end{tabular}}} \\ \hline
\textit{"I was able to easily recreate the given environment"}       & 8.5                                                                          \\ \hline
\textit{"I was able to create a new environment to my satisfaction"} & 8.7                                                                          \\ \hline
\end{tabular}
    \linebreak
    \caption{Compiled results of experimental tasks to showcase the usability of HEROES during replication tasks and creating new environments}
    \label{table:task-eval}
\end{table}
\begin{table}[!htbp]
\begin{tabular}{|l|l|}
\hline
\multicolumn{1}{|c|}{\textit{\textbf{Questions}}}                                                                                                                                                                            & \multicolumn{1}{c|}{\textit{\textbf{\begin{tabular}[c]{@{}c@{}}Average\\ Score\end{tabular}}}} \\ \hline
\textit{\begin{tabular}[c]{@{}l@{}}\underline{Usability}\\ 1. I was able to navigate easily within the simulator\\ 2. The simulator’s user interface is easy to use\\ 3. I was able to generate varied environments easily\end{tabular}} & \begin{tabular}[c]{@{}l@{}}\\ 8.5\\ 8.3\\ 8.1\end{tabular}                                   \\ \hline
\textit{\begin{tabular}[c]{@{}l@{}}\underline{Effectiveness as a Tool for Robot Training}\\ 1. HEROES provided a useful simulation environment\\ 2. HEROES provides a realistic portrayal of the\\ Post MCI environment\end{tabular}}    & \begin{tabular}[c]{@{}l@{}}\\ 7.8\\ 7.9\end{tabular}                                          \\ \hline
\end{tabular}
    \caption{User Feedback Results regarding the general usability of HEROES and its ability as a replication of Post MCI environments.}
    \label{table:usability-eval}
\end{table}

\section{Real-World Implementation and Testing}
To validate the real-world applicability of the HEROES simulation framework, we focused on the transfer of policies learned in simulation to physical robots operating in complex post-MCI terrains. Specifically, we used HEROES to generate high-fidelity 3D simulated environments and train a quadrupedal robot, the Unitree Go1, using Reinforcement Learning (RL). This section provides an in-depth overview of the training pipeline, the integration of 3D simulation data, and domain randomization techniques used.

\subsection{Leveraging 3D Simulated Data for Training}
The HEROES simulator generates rich, detailed 3D environments consisting of destructible terrain, rubble, and structural debris that dynamically interact with robotic agents. These simulated environments serve as the primary source of training data for the RL-based navigation algorithm. Each simulated environment is represented as a collection of meshes, textures, and physical interactions. The robot perceives the environment through multiple simulated sensors, including RGB cameras, depth sensors, and LiDAR, providing it with a comprehensive understanding of the 3D world. The sensory data from the simulator, particularly the depth maps and semantic segmentation, is used as input to the policy network, allowing the robot to estimate the traversability of the terrain and plan its motion accordingly.

Let \( \mathbf{x}_t = [I_t, D_t, S_t] \) represent the sensory input at time \( t \), where, \( I_t \) is the RGB image data, \( D_t \) is the depth map data, and \( S_t \) is the semantic segmentation data. This input is processed by a convolutional neural network (CNN) to extract high-level features \( \phi(\mathbf{x}_t) \), which are used by the RL policy \( \pi_\theta(a_t | \phi(\mathbf{x}_t)) \) to select the optimal action \( a_t \in A \) from the action space.

The advantage of using simulated 3D data is the ability to generate an infinite number of training scenarios. By altering the geometry, lighting, material properties, and destruction patterns in the HEROES environment, the robot is exposed to highly varied conditions, ensuring that it learns generalizable navigation strategies rather than overfitting to specific terrain layouts. Each variation introduces new challenges for the robot, such as unstable rubble, occlusions, and dynamically changing pathways.

\subsection{Domain Randomization for Sim-to-Real Transfer}
Domain randomization plays a crucial role in ensuring that policies learned in the simulated environment transfer effectively to the real world. By systematically varying key parameters in the simulation, we expose the robot to a wide range of possible conditions, increasing its robustness to the variability and uncertainty inherent in real-world environments.

In the HEROES simulation, the following parameters are randomized across training episodes:
\begin{itemize}
    \item \textbf{Terrain Properties:} The friction coefficients \( \mu \) of different terrain surfaces, including concrete, wood, and rubble, are randomized within a defined range \( \mu \in [\mu_{\min}, \mu_{\max}] \).
    \item \textbf{Sensor Noise:} Gaussian noise \( \mathcal{N}(0, \sigma^2) \) is added to the RGB images, depth maps, and LiDAR readings to simulate real-world sensor inaccuracies.
    \item \textbf{Lighting Conditions:} The intensity, direction, and color temperature of the lighting in the environment are randomized to simulate different times of day and weather conditions.
    \item \textbf{Terrain Geometry:} The arrangement and size of rubble, debris, and structural elements are procedurally generated, ensuring that the robot encounters novel layouts in each episode.
    \item \textbf{Actuator Dynamics:} The torque limits and response times of the robot's actuators are varied to simulate wear and tear or manufacturing differences in real-world robots.
\end{itemize}

Mathematically, the domain randomization can be represented as introducing variability in the parameters \( \tau \) of the environment dynamics model \( p_\tau(s_{t+1} | s_t, a_t) \). For each episode, \( \tau \) is sampled from a distribution \( \mathcal{P}_\tau \), and the robot's policy \( \pi_\theta(a_t | s_t) \) is trained to maximize the expected reward across all sampled environments:
\begin{equation}
    \max_\theta \mathbb{E}_{\tau \sim \mathcal{P}_\tau} \left[ \sum_{t=0}^{T} r_t(s_t, a_t, \tau) \right].
\end{equation}

By exposing the policy to randomized parameters, we ensure that it learns to adapt to a broad range of conditions, making it less sensitive to the specific details of the training environment and more capable of generalizing to new, unseen terrains.

\subsection{Training Procedure}
The robot's policy is trained using the Proximal Policy Optimization (PPO) algorithm, which updates the policy by minimizing the clipped surrogate loss function:
\begin{equation}
    L^{CLIP}(\theta) = \mathbb{E}_t \left[ \min(r_t(\theta) \hat{A}_t, \text{clip}(r_t(\theta), 1 - \epsilon, 1 + \epsilon) \hat{A}_t) \right],
\end{equation}
where \( r_t(\theta) \) is the probability ratio between the current and previous policies, \( \hat{A}_t \) is the advantage function, and \( \epsilon \) is a hyperparameter controlling the update size.

To accelerate the training process, we use parallelized training environments, allowing multiple instances of the HEROES simulator to run concurrently. This approach increases the diversity of experiences and helps the policy converge faster by exploring different randomizations in each simulation instance.

\subsection{Real-World Testing}
After training in the simulated environment, the learned policy was transferred to the Unitree Go1 robot for real-world testing. The robot was deployed in an outdoor rubble field with conditions designed to mimic those seen in the simulation. Key performance metrics such as traversal time, collision rate, and energy efficiency were recorded and compared to simulated performance.
One key challenge in the real-world deployment is the inevitable \emph{reality gap}—the difference between the simplified simulation model and the complexity of real-world physics. To address this, we further fine-tuned the policy on real-world data using an online adaptation strategy, where the policy is updated in real-time based on new experiences gathered in the real environment:
$   \theta \leftarrow \theta + \alpha \nabla_\theta L(\theta)$, where \( \alpha \) is the learning rate for online adaptation.
The real-world experiments demonstrated that the combination of 3D simulated data and domain randomization allowed the robot to successfully navigate highly complex and dynamic terrains. Although some degradation in performance was observed due to unmodeled real-world phenomena (e.g., unaccounted sensor noise, more chaotic terrain deformation), the overall behavior of the robot closely aligned with its simulated performance, validating the sim-to-real transfer.

The detailed evaluation in an MCI scenario is beyond the scope of this work. This work focuses on the generation/simulation of physics-inspired terrains in MCI. We have also open-sourced the code for HEROES: \href{https://github.com/Anav-117/HEROES}{https://github.com/Anav-117/HEROES}.  

% \section{Conclusions}
% We presented HEROES- a versatile simulator designed for training humans and emergency robots for such urban search and rescue operations. By using destructible environments, we are able to generate scenarios where hazardous terrain presents a serious challenge to navigation. Training in such environments can help better ascertain drawbacks in existing methods of robot navigation when confronted with unstable terrain not easily replicable in the real world. Furthermore, HEROES provides a logistically feasible method to simulate a wide variety of post-MCI scenarios that cannot be performed in real-world training. 
% In the future, we hope to extend the capabilities of the simulator to allow users to easily train multiple robots in a variety of scenarios. We also aim to expand the simulator's capabilities to allow users to train heterogeneous teams of robots covering MCIs beyond simple destroyed urban environments for uneven terrain navigation. 
\section{Conclusion}

We introduced HEROES, a versatile simulation platform for training emergency robots and humans in urban search and rescue operations. By simulating destructible environments, HEROES replicates complex post-Mass Casualty Incident (MCI) terrains that pose significant challenges to navigation. These scenarios, difficult to replicate in real-world settings, allow for a thorough evaluation of robotic navigation strategies under hazardous conditions. HEROES also provides a scalable and cost-efficient framework for simulating diverse MCI environments, advancing preparedness for real-world disaster response.

While HEROES represents a substantial improvement in simulating complex environments, certain limitations remain. Despite the use of high-fidelity physics engines, a gap exists between the simulated and real-world conditions, which affects the transferability of learned behaviors. Additionally, the detailed environments require considerable computational resources, potentially limiting accessibility for some users. The current version of the simulator also lacks full representation of nuanced human and robot behaviors under extreme stress, reducing the realism of some training scenarios.

To address these limitations, future work will extend HEROES to support multi-robot training in a broader range of scenarios, including heterogeneous robot teams and non-urban MCI environments. These advancements will further enhance HEROES' role as a critical tool for improving the safety and efficiency of search and rescue operations in unpredictable and complex disaster settings.

\pagebreak

%% Bibliography
\bibliographystyle{IEEEtran}
\bibliography{root}

\end{document}